%% file: main.tex
\documentclass[letterpaper, 10 pt, conference]{ieeeconf}  % Comment this line out if you need a4paper
\usepackage{url}
\usepackage{xcolor,colortbl}
\usepackage{amsmath,amsfonts,amssymb,wasysym}
\usepackage{verbatim}
\usepackage{mathrsfs}
\usepackage[noend]{algorithmic}
\let\chapter\section
\usepackage{wrapfig}
\usepackage{graphicx} 
\usepackage{xspace}
\usepackage{capt-of}% or \usepackage{caption}
\usepackage[ruled,linesnumbered]{algorithm2e}

\usepackage{multirow,multicol}
\usepackage{color}
\usepackage{soul}
\SetKwComment{Comment}{$\triangleright$\ }{}
\usepackage{cite}

\IEEEoverridecommandlockouts                              % This command is only needed if 
\title{\LARGE \bf
A Probabilistic Model for Planar Sliding of Objects with Unknown Material Properties: Identification and Robust Planning
}

\author{
Changkyu Song and Abdeslam Boularias$^{1}$
\thanks{$^{1}$The authors are with the Department of Computer Science of Rutgers University, Piscataway, New Jersey 08854, USA.
        {\tt\footnotesize \{cs1080, ab1544\}@cs.rutgers.edu}.
        This work was supported by NSF awards 1734492, 1723869 and 1846043.}
}

\begin{document}

\maketitle

\thispagestyle{empty}
\pagestyle{empty}

%%%%%%%%%%%%%%%%%%%%%%%%%%%%%%%%%%%%%%%%%%%%%%%%%%%%%%%%%%%%%%%%%%%%%%%%%%%%%%%%
\begin{abstract}
This paper introduces a new technique for learning probabilistic models of mass and friction distributions of unknown objects, and performing robust  sliding actions by using the learned models. The proposed method is executed in two consecutive phases. In the exploration phase, a table-top  object is poked by a robot from different angles. The observed motions of the object are compared against simulated motions with various hypothesized mass and friction models. The simulation-to-reality gap is then differentiated with respect to the unknown mass and friction parameters, and the analytically computed gradient is used to optimize those parameters. Since it is difficult to disentangle the mass from the friction coefficients in low-data and quasi-static motion regimes, our approach retains a set of locally optimal pairs of mass and friction models. A probability distribution on the models is computed based on the relative accuracy of each pair of models. In the exploitation phase, a probabilistic planner is used to select a goal configuration and waypoints that are stable with a high confidence. The proposed technique is evaluated on real objects and using a real manipulator. The results show that this technique can not only identify accurately mass and friction coefficients of non-uniform heterogeneous objects, but can also be used to successfully slide an unknown object to the edge of a table and pick it up from there, without any human assistance or feedback.
\end{abstract}

\input{introduction}

\input{related}

\input{setup}

\input{proposed}

\input{experiments}

\bibliographystyle{IEEEtran}
\bibliography{bibliography.bib}
\end{document}

%% file: introduction.tex
\section{Introduction}
Pre-grasp sliding manipulation of objects is a useful skill that is necessary when an object is too thin relative to the size of a robotic hand to be directly grasped from a flat surface. This skill is also useful for the execution of tool-use grasps where the orientation of an object needs to be modified before picking up the object. For example, a pre-grasp sliding manipulation approach that was recently developed~\cite{Kaiyu2019} consists in performing a sequence of non-prehensile actions such as side-pushing and top-sliding to relocate an object to the edge of a table so that part of the object sticks out of the table. The object could then be grasped by a robotic hand without colliding with the table. 

Most existing methods for pre-grasp sliding manipulation require the existence of predefined geometric and mechanical models of the target object~\cite{Dogar2010PushgraspingWD,6631288,Dogar_2012_7076,King2013PregraspMA,Dogar2011AFF,Dogar2012}. Such models are necessary for selecting appropriate pushing forces and ensuring the balance of the object when it is moved to the edge of the support surface. In this work, we relax this assumption and consider the general problem of side-pushing an unknown object toward a stable goal configuration.

The proposed approach consists in first identifying the mass distribution of the target object, and its coefficients of friction with the support surface, from observed motions and rotations of the object. While there are many techniques for identifying mechanical properties of objects from data~\cite{Mason86,doi:10.1177/027836499601500602,23847993b652419a91558fd1f03bbec3,doi:10.1177/027836499601500603,Dogar2010PushgraspingWD,DBLP:conf/icra/ZhouPBM16,DBLP:journals/ijrr/ZhouHM19,DBLP:journals/ijrr/ZhouMPB18,Yoshikawa1991IndentificationOT}, existing works identify only the friction parameters, and assume that the mass distribution is known. 
Moreover, several existing methods assume that the bottom surface of the object has a uniform homogenous friction everywhere. In this work, we consider a grid-based representation of the target object. The object is modeled as a large set of small cuboids
that are attached to each other. The mass and coefficient of friction of each cuboid can be different from the others. The analytical gradient of the object's motion is then used to learn the mechanical properties of each cuboid.

\begin{figure}[t]
    \centering
        \includegraphics[width=1\linewidth]{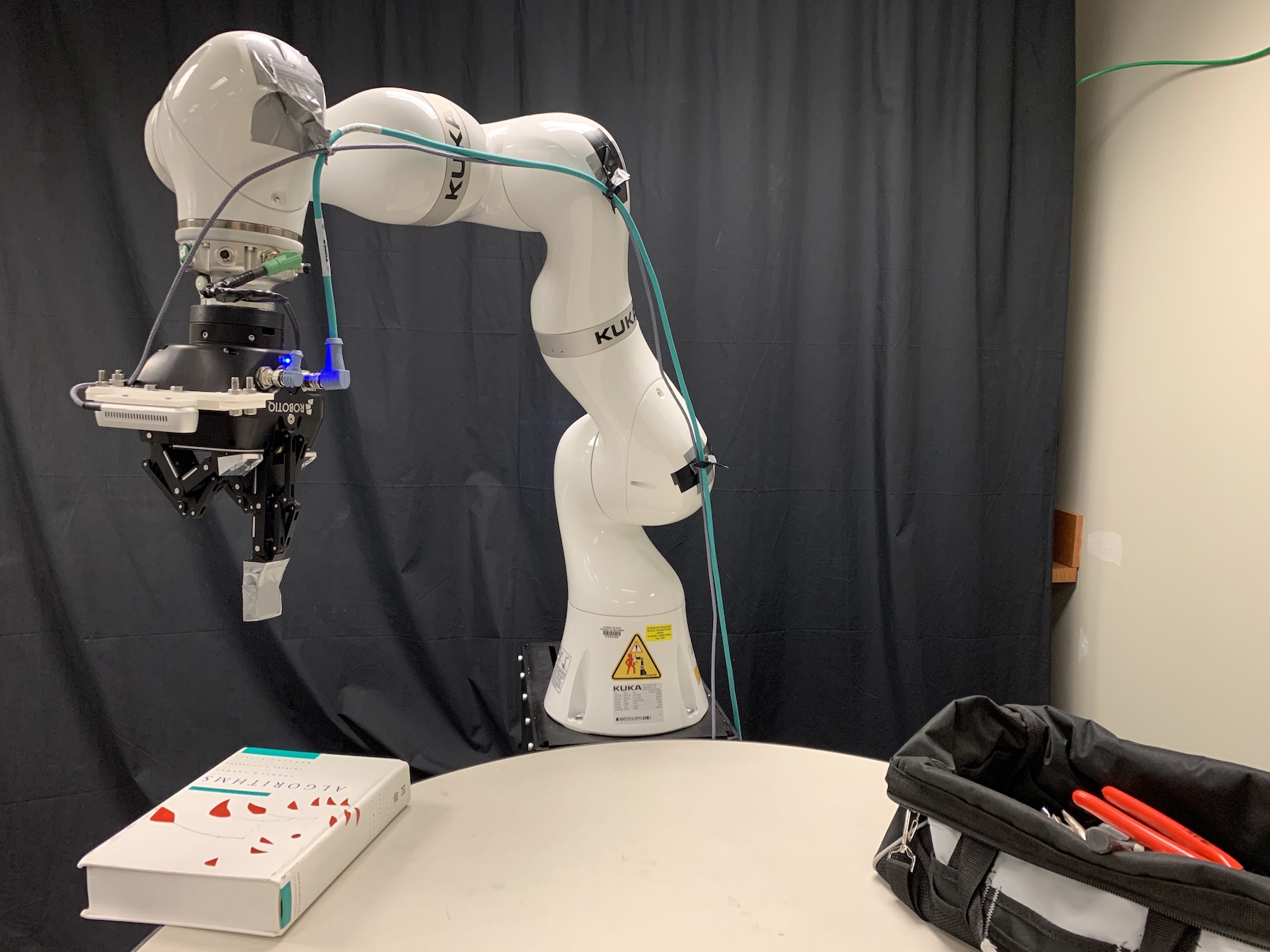}
    \caption{Robotic setup used in the experiments. In this scenario, the robot pushes a book into a desired final pose through consecutive pushes. The choice of contact points and pushing directions is non-trivial as it depends on the mass distribution and the friction coefficients of different parts of the object, which are unknown {\it a priori}.}
    \label{fig:robot}
\end{figure}

The problem of identifying the mass distribution and friction coefficients of an unknown object is under-determined. From the small number of data points collected by the robot by pushing the object, there is a large subspace of possible mass and friction values that can be identified and that explain the observed motions equally well. This problem is particularly pronounced in quasi-static motions, where the mass of the object cannot easily be disentangled from its friction with the support surface. To solve this problem, the full spectrum of possible mass and friction values should be retained. Therefore, our method returns a probability distribution of the mass and friction parameters that can be used to deal with this problem of uncertainty. In a nutshell, we perform several gradient-descent searches starting from different random initial points, and retain all the local minima. A soft-max function is then applied on the values of the local minima to obtain a probability distribution on possible mass and friction values. 

The obtained distribution is used, in a second phase, for robust probabilistic planning.
Planning here refers to the problem of finding a stable goal configuration and a sequence of side-pushing actions that displace the object to the goal while preserving its balance. Probabilistic planning consists in searching for actions in a {\it multiverse} simulation~\cite{ISRR2019}, wherein next states are sampled from various models of mass and friction of the object, with different probabilities. The sequence of actions that has the highest success probability is then selected and executed. 
The proposed technique for probabilistic model identification and manipulation planning is evaluated both in simulation and on real objects and robot. 
The results clearly confirm the advantage of using this technique to safely manipulate unknown objects.

%% file: related.tex
\section{Related Work}
Past works have explored the {\it mechanics of pushing}~\cite{Mason86,doi:10.1177/027836499601500602,23847993b652419a91558fd1f03bbec3,doi:10.1177/027836499601500603,Dogar2010PushgraspingWD,DBLP:conf/icra/ZhouPBM16,DBLP:journals/ijrr/ZhouHM19,DBLP:journals/ijrr/ZhouMPB18,Yoshikawa1991IndentificationOT}. 
For example, Mason~\cite{Mason86} studied the rotation and translation of an object pushed at a single contact point. Lynch and Mason~\cite{doi:10.1177/027836499601500602} also proposed an algorithm for stable pushing with a robotic end-effector. Yoshikawa and Kurisu~\cite{Yoshikawa1991IndentificationOT} showed that a regression method can learn pivot points of a pushed object. A similar setup was considered in~\cite{23847993b652419a91558fd1f03bbec3} with a constraint to ensure positive friction coefficients. The {\it limit surface} is an important concept in these works, it is defined as a convex set of all friction wrenches that can be applied on an object, and it is generally approximated as an ellipsoid~\cite{doi:10.1177/027836499601500603}, or a higher-order polynomial~\cite{DBLP:conf/icra/ZhouPBM16,DBLP:journals/ijrr/ZhouHM19,DBLP:journals/ijrr/ZhouMPB18,Dogar2010PushgraspingWD}.
In contrast to our method, these works identify only the friction parameters, and assume that the mass distribution is known. Moreover, the present work focuses on safe planning when the identified mass and friction models are uncertain due to the small number of actions used to identify the models in a quasi-static regime, in contrast to prior works~\cite{ChangkyuRSS2020,L4DC2020Changkyu,Belbute-Peres2017} that consider deterministic models.

Pre-grasp sliding was also considered in prior works with deterministic pre-defined models~\cite{Kaiyu2019,PackingICRA2019,ChangkyuIROS2019}, black-box model identification~\cite{ShaojunIJCAI2018}, or model-free reinforcement learning~\cite{BoulariasBS15}.

\begin{comment}
Several \emph{natively differentiable physics engines} have been proposed recently~\cite{DegraveHDW16}.
For example, a combination of a learned and a differentiable engine was used to learn to predict motions of planar objects~\cite{DBLP:journals/corr/abs-1710-04102}. 
Differentiable physics engines were also used in a planning algorithm to manipulate tools~\cite{18-toussaint-RSS}. A differentiable contact model was used in~\cite{Mordatch:2012} to allow for optimization of several  locomotion and manipulation tasks. A differentiable physical simulation, formulated as a Linear Complementary Problem (LCP), was also recently proposed~\cite{Belbute-Peres2017}. 
\end{comment}

%% file: setup.tex
\section{Problem Setup and Notations}
\label{sec:setup}
We revisit the problem of pre-grasp sliding manipulation presented in~\cite{Kaiyu2019}. In this problem, the robot's task consists in grasping a rigid object placed on a table. The object cannot be easily grasped, so the robot slides it to the table's edge where part of it sticks out of the table and becomes exposed. The robot can then easily grasp it from there. In this work, we consider the same problem, but for objects with unknown shape and material properties. The problem then consists in 1) selecting pushing actions to explore the object and gather data points about it, 2) inferring the mass and friction distributions of the object from the observed motions, 3) selecting a stable goal configuration $x^d_T$ from which the object can be grasped, and 4) planning a sequence of side-pushing actions that 
displace the object from its current pose $x_0$ to the desired final pose $x^d_T$. 
 
 The object is represented as a set of small cuboids. The object is thus divided into a large number of connected cells $1,2,\dots,n$, using a regular grid structure, wherein each cell $i$ has its own local mass and coefficient of friction. 
The object's pose $x_t$ at time $t\in [0,T]$ is given as a vector $x_t = [p^1_{x,t},p^1_{y,t},\theta^1_t,\dots, p^n_{x,t},p^n_{y,t},\theta^n_t]^T$, where $(p^i_{x,t},p^i_{y,t})$ is the $i^{th}$ cell's 2D position on the surface, and $\theta^i_t$ is its angle of rotation. The object's generalized velocity at time $t$ is denoted as $\dot{x}_t = [\dot{p}^i_{x,t},\dot{p}^i_{y,t},\dot{\theta}^i_t]_{i=0}^n$.
The object's mass matrix $\mathcal{M}$ is a diagonal $3n\times3n$ matrix, where the diagonal is $[\mathcal{I}_1,\mathcal{M}_1,\mathcal{M}_1,\mathcal{I}_2,\mathcal{M}_2,\mathcal{M}_2,\dots, \mathcal{I}_n,\mathcal{M}_n,\mathcal{M}_n]$, $\mathcal{I}_i$ is the moment of inertia of the $i^{th}$ cell of the object, and $\mathcal{M}_i$ is its mass. $\mathcal{I}_i = \frac{1}{6} \mathcal{M}_i w^2$ where $w$ is the width of a cuboid.
$\mu$ is a $3n\times 3n$ diagonal matrix, where the diagonal is $[\mu_1,\mu_1,\mu_1,\mu_2,\mu_2,\mu_2,\dots,\mu_n,\mu_n,\mu_n]$. $\mu_i$ is the coefficient of friction between the $i^{th}$ cell of the object and the surface.

An external generalized force denoted by $F$ is an $1\times 3n$ vector $[f^1_{x},f^1_{y},\tau^1,\dots, f^n_{x},f^n_{y},\tau^n]^T$, where $[f^i_{x},f^i_{y}]$ and $\tau^n$ are respectively the force and torque applied on cell $i$. External forces are generated from the contact between the object and a fingertip of the robotic hand used to push the object. We assume that at any given time $t$, at most one cell of the object is in contact with the fingertip.

A ground-truth trajectory $\mathcal T^g$ is a state-action sequence $(x^g_0,\dot{x}^g_0,F_{0}, \dots, x^g_{T-1},\dot{x}^g_{T-1}, F_{T-1}, x^g_{T},\dot{x}^g_{T})$, wherein $(x^g_t,\dot{x}^g_{t})$ is the observed pose and velocity of the pushed object, and $F_{t}$ is the external force applied at time $t$, as defined above.
A corresponding simulated trajectory $\mathcal T$ is obtained by starting at the same initial state
$\hat{x}_0$ in the
corresponding real trajectory, i.e., $\hat{x}_0 = x_0$, and applying
the same control sequence $(F_0, F_1, \dots, F_{T-1})$. Thus,
the simulated trajectory $\mathcal T$ results in a state-action sequence $(x^g_0,\dot{x}^g_0,F_{0},x_1,\dot{x}_1,F_{1}, \dots, x_{T-1},\dot{x}_{T-1}, F_{T-1}, x_{T},\dot{x}_{T})$,
where $x_{t+1} = x_{t} + \dot{x}_{t}dt$ is the predicted next pose. 
Velocity $\dot{x}_{t}$ is a vector corresponding to translation  and angular velocities in the plane for each of the $n$ cells, it is predicted in simulation as
$\dot{x}_{t+1} = V(x_{t},\dot{x}_{t}, F_t, \mathcal M, \mu)$.
%In other words, the state of the simulation is reset at each time-step to the real observed state. %This helps so that the real-world trajectory $\tau^*$ and the simulated one $\tau$ do not diverge significantly one from the other.
The goal is to identify mass distribution $\mathcal M$ and friction map $\mu$ that result in simulated trajectories that are as close as possible to the real observed ones.  Therefore, the objective is to solve the following optimization problem,
{\small
\begin{eqnarray}
(\mathcal M^*, \mu^*) &=& \arg \min_{\mathcal M, \mu} loss(\mathcal M, \mu),  \label{simulationError} \\
loss(\mathcal M, \mu)\hspace{-0.2cm} &\stackrel{def}{=}& \hspace{-0.2cm}  \sum_{t=0}^{T-2}\|
\big(x^g_{t+1} +  V(x^g_{t},\dot{x}^g_{t}, F_t, \mathcal M, \mu)dt \big) - x^g_{t+2} 
\|_2. \nonumber
\end{eqnarray}
}
Since $x_{t}$ is a vector containing all cells' positions, the loss is the sum of distances between each cell's ground-truth pose and its predicted pose, which is equivalent to the average distance (ADD) metric as proposed in~\cite{Hinterstoisser2013}.

\section{Deterministic Mass and Friction Identification}
In%~\cite{ChangkyuL4DC2020},
~\cite{ChangkyuRSS2020},
it has been shown that the gradient of the loss function in Equation~\ref{simulationError} with respect to mass parameters $\mathcal M$ and friction map $\mu$ can be computed analytically, and used to search for ground-truth values $(\mathcal M^*, \mu^*)$ by using the gradient-descent method. The algorithm proposed in%~\cite{ChangkyuL4DC2020} 
~\cite{ChangkyuRSS2020},
consists in initializing $(\mathcal M, \mu)$ with arbitrary positive values, and repeatedly using $(\mathcal M, \mu)$ to generate a simulated trajectory $(x^g_0,\dot{x}^g_0,F_{0},x_1,\dot{x}_1,F_{1}, \dots, x_{T-1},\dot{x}_{T-1}, F_{T-1}, x_{T},\dot{x}_{T})$, which will then be used to update $(\mathcal M, \mu)$ as follows:
\begin{flalign}
%\mu \leftarrow \mu + \sqrt{\alpha_{\textrm{rate}}} \sum_{t=0}^{T-2} Diag\big(x_{t+2} - x^g_{t+2}\big),\nonumber\\
%\mathcal{M} \leftarrow \mathcal{M} + \alpha_{\textrm{rate}} \sum_{t=0}^{T-2}Diag\big(x_{t+2} - x^g_{t+2}\big),
%&W^t_\mu = X \mathcal{J}^T_f (\dot{x}_t)\mathcal{M}\nonumber\\
%&W^t_\mathcal{M} = X\Big(\mathcal{J}^T_f(\dot{x}_ t)\mu -D(\dot{x}_{t+1} - \dot{x}_t)\Big)\nonumber\\
&\mu \leftarrow \mu - \alpha_{\textrm{rate}} \sum_{t=0}^{T-2} Diag\big(x_{t+1} - x^g_{t+1}\big)W^t_\mu,\nonumber\\
&\mathcal{M} \leftarrow \mathcal{M} - \alpha_{\textrm{rate}} \sum_{t=0}^{T-2}Diag\big(x_{t+1} - x^g_{t+1}\big)W^t_\mathcal{M},
\label{updateMassFriction}
\end{flalign}
wherein $\alpha_{\textrm{rate}}$ is a predefined step-size, and $W^t_\mu$ and $W^t_\mathcal{M}$ are weighting matrices provided in~\cite{ChangkyuRSS2020} and computed based on the manipulated object's geometry, current position and velocity. While this approach is shown to return an accurate map of the frictional forces on the surface of a non-uniform object, it cannot disentangle the friction from the mass if all the data points correspond to quasi-static motions. In other terms, the product mass$\times$friction of the identified mass and friction of each cell in the object is close to its ground-truth value, and it is accurate enough to predict the translation and rotation of the object under various pushing actions. However, it cannot tell which part of the product mass$\times$friction is mass and which part is friction. For example, if the ground-truth friction force at some cell of the object is $\mathcal M^*_i \times \mu^*_i = 10 g \textrm{ N}$, where $g = 9.8$, then the update equations~\ref{updateMassFriction} can typically identify a value for $\mathcal M_i$ and for $\mu_i$ such that $\mathcal M_i \times \mu_i \approx \mathcal M^*_i \times \mu^*_i = 10 g \textrm{ N}$ and so that the translation and rotation of the object can be predicted accurately in quasi-static motions on a flat surface. But the identified individual values for mass and friction can be very far from the ground-truth values. For example, any pair of values from the set $\{(1kg,98\frac{\textrm{N}}{kg}), (98kg,1\frac{\textrm{N}}{kg}), (2kg,49\frac{\textrm{N}}{kg}), (20kg,4.9\frac{\textrm{N}}{kg}), \dots \}$ can be returned by the deterministic identification algorithm proposed in~\cite{ChangkyuRSS2020}.

%Given an identified model $\theta^*$, our objective is to identify a sequence of pushing forces $(u^*_i)_{i=t}^{H} \stackrel{def}{=}U_t^*$ that tracks a trajectory $(x^*_i)_{i=t}^{H} \stackrel{def}{=} X_t^*$ such that:
%\begin{eqnarray}
%U_t^* \stackrel{def}{=} \arg \min_{U_t} J(x_t,U_t,X^*_t,\theta^*)
%\label{optimalPolicy}\\
%J(x_t,U_t,X^*_t,\theta^*) \stackrel{def}{=} \sum_{i=t}^{H} ( \|u_i\|_2 + \|x_i-x^*_i \|_2 ),\\ x_i=F(x_{i-1}, u_i, \theta^*), i\in\{t+1,\dots, H\}
%\label{costFunction}
%\end{eqnarray}

%% file: proposed.tex
\section{Proposed Approach}
While the exact value of the mass of each cell of the object is relatively irrelevant in quasi-static motions, that is not the case when the object is pushed to the edge of the table and where one needs to predict if it remains stable there or if it drops, which is necessary for pre-grasp sliding manipulation. 
In the following we propose a solution to this problem by first presenting a probabilistic model identification approach that returns a distribution on joint mass and frictions maps, instead of single point estimates. We then show that the probabilistic model can be used for robustly selecting stable goal configurations and planning pre-grasp sliding manipulation actions. 

\subsection{Probabilistic Mass and Friction Model}
We generalize the deterministic model presented in Section~\ref{sec:setup} into a probabilistic one. The probabilistic model is given by a set of mass and friction maps denoted by $\Theta = \{ (\mathcal M^{(1)}, \mu^{(1)}) , (\mathcal M^{(2)}, \mu^{(2)}) , \dots (\mathcal M^{(K)}, \mu^{(K)}) \}$ and a probability measure $P: \Theta \rightarrow [0,1]$. Each $(\mathcal M^{(k)}, \mu^{(k)})$ corresponds to one of $K$ sampled models, and is defined as explained in Section~\ref{sec:setup}. Thus, $\mathcal M^{(k)}$ is a diagonal matrix, where the diagonal corresponds to the moment of inertia and the mass of each cell of the object, and 
$\mu^{(k)}$ is a diagonal matrix, where the diagonal corresponds to the coefficients of friction between each of the $n$ cells of the object and the support surface. We assume that: $\forall k\in \{1,\dots, K\}, \forall i\in \{1,\dots, n\}: \mu^{(k)}_i\in[0,\mu_{max}]$, and $\mathcal M^{(k)}_i\in[0,\mathcal M_{max}]$ where $\mu_{max}$ and $\mathcal M_{max}$ are given upper bounds. We will explain in the following how models in $\Theta$ are sampled and how their probabilities are computed.

\subsection{Inference of Probabilistic Models}
The proposed inference algorithm consists in collecting a real-world trajectory  $ (x^g_{t},\dot{x}^g_{t},F_{t})_{t=0}^{T}$ of ground-truth object poses and velocities that result from a sequence of selected exploratory pushing forces $(F_{t})_{t=0}^{T}$. Different models $(\mathcal{M}^{(k)}, \mu^{(k)})$ of mass and friction distributions are then returned by the algorithm, with different probabilities. Each one of the models is obtained by upper-bounding the mass and the friction with a different limit, and using the gradient-descent method to search for values of mass and friction that minimize the simulation-to-reality gap. 

The main steps of this search process are explained in Algorithm~\ref{identificationAlgo}.
The set $\Theta$ of sampled models is initially empty. The first pushing action is selected arbitrarily as no prior information about the object's mass or friction distributions is available. The magnitudes of all the pushing actions are small to avoid moving the object out of the robot's workspace. At each iteration of the algorithm, an action is selected by calling the subroutine \texttt{ActionSelection} presented in Algorithm~\ref{actionSelection}, which will be explained in the next subsection. The returned action $F_T$ is applied on the object. The next pose and velocity of the object
$(x^g_{T+1},\dot{x}^g_{T+1})$ is extracted from recorded depth images after removing the background. 

The inner loop of the algorithm (lines 7-27) consists in setting different upper bounds for mass and friction and optimizing $(\mathcal{M}^{(k)}, \mu^{(k)})$ to produce simulated trajectories of the pushed object that are as close as possible to the recorded ground-truth.
In the first half of the total number of iterations $K$, the upper bound of the mass per cell is fixed to $\mathcal M_{max}\in \mathbb{R}$ whereas the upper bound of the friction is gradually increased at each iteration. In the second half, the upper bound of the friction is fixed to $\mu_{max}\in \mathbb{R}$ whereas the upper bound of the mass per cell is gradually increased at each iteration. Other combinations of mass and friction upper bounds can also be considered as well. The provided upper limits are the same for every cell in the object, whereas the mass and friction maps learned by the algorithm are highly heterogenous, as will be shown in the experiments.

At each iteration $k$ of the inner loop, the state of the object in simulation $(x_0,\dot{x}_0)$ is reset to the ground-truth initial state $(x^g_0,\dot{x}^g_0)$ (line 13). Both the mass matrix $\mathcal{M}^{(k)}$ and the friction matrix and $\mu^{(k)}$ are initialized to half their maximum values (lines 14-15). In the next steps (lines 17-25), $\mathcal{M}^{(k)}$ and $\mu^{(k)}$ are iteratively used to predict in simulation the trajectory of the manipulated object under the same forces $\{F_t\}_{t=0}^{T}$, and updated so that the predicted pose $x_{t+1}$ is as close as possible to the observed ground-truth $x^g_{t+1}$. The sum of the differences over all time-steps is denoted by $loss^{(k)}$. The mass and friction matrices are updated by using the stochastic gradient of the loss (lines 21-22), wherein $Diag\big(x_{t+1} - x^g_{t+1}\big)$ is a diagonal matrix that contains $\big(x_{t+1} - x^g_{t+1}\big)$ in the main diagonal, in accordance with the definitions of $\mathcal{M}$ and $\mu$ provided in Section~\ref{sec:setup}. The mass and friction are increased or decreased depending on the signs in the error vector $\big(x_{t+1} - x^g_{t+1}\big)$, which corresponds to the reality gap.

Finally, the models identified from the full sequence of pushing actions are weighed based on their respective accuracies, using a soft-max function (line 30). The soft-max operator returns a probability distribution over competing models of mass and friction distributions. Each of these models corresponds to a local minimum of the reality-to-simulation loss, obtained by gradient-descent. Despite the fact that most of these models achieve a very small loss in predicting the motion of the object on a flat surface, they vary significantly in the way they attribute the observed motion to the mass or to the friction coefficients. While this difference is not consequential on a flat surface, it is important for predicting the balance of the object when it is pushed all the way to the edge of the support surface. 

\begin{algorithm}[h]
{\small
     \SetAlgoLined
     \KwIn{Learning rate $\alpha_{\textrm{rate}}$; Total number of pushing actions $nb\_actions$; Upper bounds $\mu_{max}$ and $\mathcal M_{max}$\;}
\KwOut{A set $\Theta$ of mass and friction maps, and a corresponding probability distribution $P$\;}

Sample a random contact point from the object's contour, and use the robot's fingertip to push the object from the sampled point with a small random force $F_0$; Record the observed new state $(x^g_1,\dot{x}^g_1)$\;
$\Theta \leftarrow \emptyset$\;
\For{$T = 1,nb\_actions$}
{
    $F_T\leftarrow$ \texttt{ActionSelection}($\Theta, \{F_0,\dots, F_{T-1}\}$)\;
    Apply $F_T$; Record the observed new state $(x^g_{T+1},\dot{x}^g_{T+1})$\;
    $\Theta \leftarrow \emptyset$\;
    \For{$k = 1, K$}
    {
        \eIf{$k \leq \frac{K}{2}$}
        {${\mu}_{max}^{(k)}\leftarrow \frac{2k}{K}{\mu}_{max}$;
        ${\mathcal M}_{max}^{(k)}\leftarrow {\mathcal M}_{max}$\;
        }
        {${\mu}_{max}^{(k)}\leftarrow {\mu}_{max}$;
        ${\mathcal M}_{max}^{(k)}\leftarrow (\frac{2k}{K}-1) {\mathcal M}_{max}$\;
        }
        $(x_0,\dot{x}_0) \leftarrow (x^g_0,\dot{x}^g_0)$; $x_1 \leftarrow x^g_1$\;
        $\mathcal M^{(k)} \leftarrow \frac{1}{2} I{\mathcal M}_{max}^{(k)}$\;
        $\mu^{(k)} \leftarrow \frac{1}{2} I{\mu}_{max}^{(k)}$\;
        $loss^{(k)}\leftarrow 0$\;
        \For{$t = 0, T-2$}
        {
           $\dot{x}_{t+1} \leftarrow V(x_{t},\dot{x}_{t}, F_t, \mathcal M^{(k)}, \mu^{(k)}) $ \Comment*[r]{\tiny \textcolor{blue}{Predicting velocity}}
           $x_{t+2} \leftarrow x_{t+1} + \dot{x}_{t+1}dt$\Comment*[r]{\tiny \textcolor{blue}{Predicting next pose}}
           $loss^{(k)}\leftarrow loss^{(k)} + \| x_{t+2} - x^g_{t+2} \|_2$\;
         $\mu^{(k)} \leftarrow \mu^{(k)} - \alpha_{\textrm{rate}} Diag\big(x_{t+1} - x^g_{t+1}\big) W^t_\mu $\;
          $\mathcal{M}^{(k)} \leftarrow \mathcal{M}^{(k)} - \alpha_{\textrm{rate}} Diag\big(x_{t+1} - x^g_{t+1}\big)W^t_\mathcal{M}$\;
          $\mu^{(k)} \leftarrow \arg\min_{\mu'\in[\mathbf{0}, I \mu_{max}^{(k)}]} \|\mu'-\mu^{(k)}\|_\infty$\;
          $\mathcal{M}^{(k)} \leftarrow \arg\min_{\mathcal{M}'\in]\mathbf{0}, I{\mathcal M}_{max}^{(k)}]} \|\mathcal{M}'-\mathcal{M}^{(k)}\|_\infty$
            \Comment*[r]{\tiny \textcolor{blue}{Projecting the gradients}}
        }
        $\Theta \leftarrow \Theta \cup \{(\mathcal{M}^{(k)}, \mu^{(k)})\}$\;
    }
}
    \For{$k = 1, K$}
    {
    $P(\mathcal{M}^{(k)}, \mu^{(k)}) = \frac{e^{-\frac{1}{\tau} loss^{(k)}}}{\sum_{i=1}^{K} e^{-\frac{1}{\tau} loss^{(i)}}}$\;
    }
\caption{Inference of Probabilistic Mass and Friction Models with Differentiable Physics}
\label{identificationAlgo}
}
\end{algorithm}

\begin{algorithm}[h]
    Create a set $\mathcal S$ of random candidate forces applied on random contact points in the object's contour\;
    \If{\texttt{SelectionMode = random}}
    {$F_T \leftarrow{ \texttt{RandomSample}(\mathcal S) }$ \;}
    \If{\texttt{SelectionMode = mostDifferent}}
    {$F_T \leftarrow{ \arg\max_{F\in \mathcal S}\min_{t\in[0,T-1]}\|F_t-F\|_2 }$ \;}
    \If{\texttt{SelectionMode = mostDistinctive}}
    {
        $D_{max}\leftarrow{0}$\;
        \ForEach{$F\in \mathcal S$}
        {
            $D\leftarrow{0}$\;
            \ForEach{$(\mathcal{M}^{(i)}, \mu^{(i)}), (\mathcal{M}^{(j)}, \mu^{(j)}) \in \Theta$}
            {
                $D\leftarrow D+ \| V(x_{0},\dot{x}_{0}, F, \mathcal M^{(i)}, \mu^{(i)}) $\\$- V(x_{0},\dot{x}_{0}, F, \mathcal M^{(j)}, \mu^{(j)}) \|_2$\;    
            }
            \If{$D > D_{max}$}
            {$D_{max}\leftarrow{D}$; $F_T\leftarrow{F}$\;}
        }
    }
    \KwRet{$F_T$}\;
        \caption{\texttt{ActionSelection}($\Theta, \{F_0,\dots, F_{T-1}\}$)}
    \label{actionSelection}
\end{algorithm}

  \begin{algorithm}[h]
{  \small
     \SetAlgoLined
\KwIn{A set $\Theta$ of mass and friction maps, and a corresponding probability distribution $P$\;}
\KwOut{A sequence of pushing forces $(F_t)_{t=0}^{T-1}$ ; }
Find a set $\mathcal X_{waypoint}$ of waypoints by calling $RRT^*(x_0, x_T^d)$\;
$\dot{x}_0 \leftarrow \mathbf{0}; t \leftarrow 0 $\;
\While{$ \mathcal X_{waypoint} \neq \emptyset $}
{
$x_{target} = \arg\min_{x\in \mathcal X_{waypoint}} \|x- x_t\|_2$\;
\Repeat{$\min_{x\in \mathcal X_{waypoint}} \|x- x_t\|_2 \leq \epsilon$}{
   $gap_{min}\leftarrow \|x_0 - x_T^d\|_2$\;
   \Repeat{timeout}{
   {\footnotesize Select cell $i$ from the contour of the object\;}

      {\footnotesize Choose the direction of the force $(f^i_{x},f^i_{y})$ as the opposite of the object' surface normal at cell $i$;}
   $F^i \leftarrow [0,0,0,\dots,f^i_{x},f^i_{y},0,\dots, 0,0,0]$\;
   $gap \leftarrow{0}$\;
   \ForEach{$(\mathcal{M}^{(k)}, \mu^{(k)}) \in \Theta$}
    {
      $gap \leftarrow gap + P(\mathcal M^{(k)}, \mu^{(k)}) \| x_{t} + \dot{x}_{t}dt + V(x_{t},\dot{x}_{t}, F^i, \mathcal M^{(k)}, \mu^{(k)})dt - x_{target} \|_2 $\;
    }
   \If{$gap \leq gap_{min}$}{$gap_{min} \leftarrow gap $; $F_t \leftarrow [0,0,0,\dots,f^{i}_{x},f^{i}_{y},0,\dots, 0,0,0] $\;}
   %\Comment*[r]{\tiny \textcolor{blue}{Optimizing the contact point}}
   }
   $\dot{x}_{t+1} \leftarrow V(x_{t},\dot{x}_{t}, F_t, \mathcal M, \mu)$; ${x}_{t+1} \leftarrow {x}_t + {\dot{x}_t}dt$\;
$t\leftarrow t+1$\;
}
{\small
$\mathcal X_{waypoint} \leftarrow \mathcal X_{waypoint} \backslash  \{\arg\min_{x\in \mathcal X_{waypoint}} \|x- x_t\|_2\}$}\;
}
\caption{Robust Planning with Inferred Models}
\label{policyAlgo}
}
\end{algorithm}

\subsection{Active Exploration}
The data used to identify the material properties of the manipulated object is actively collected by the robot. Algorithm~\ref{actionSelection} summarizes three different strategies that can be followed for selecting the pushing forces. First, a finite set $\mathcal S$ of candidate forces is generated by randomly sampling contact points from the object's contour, along with pushing directions and force magnitudes. 
The simplest strategy consists in returning a randomly selected force $F_T\in \mathcal S$. Another strategy selects the action that is most different from the previously executed forces $\{F_0,\dots, F_{T-1}\}$, to increase the diversity of the data and to poke different parts of the object. The third strategy consists in simulating the effect of each force $F\in \mathcal S$ on each possible (mass,friction) model from the set of identified models $\Theta$. The action that leads to the highest disagreement between the models is then selected.  

\subsection{Robust Planning with Inferred Models}
After a set $\Theta$ of possible (mass, friction) models is identified in the exploration phase by using Algorithm~\ref{identificationAlgo}, Algorithm~\ref{policyAlgo} is called
to plan a new sequence of pushing actions that displace the object into a desired final pose $x^{d}_{T}$. We start by generating a sequence of intermediate poses, i.e., {\it waypoints}, by using the rapidly-exploring random tree ($RRT^*$) algorithm. 
Our algorithm then generates a sequence of pushing forces $(F_t)_{t=0}^{T-1}$ that track the waypoints. At each step in the sequence, a force $F_t$ is selected and optimized so that, when executed, the resulting pose of the object is as close as possible to the nearest waypoint. There are several heuristics that can be used to select a force $F_t$ that moves the object to the nearest waypoint $x_{target}$ at time-step $t$. In this work, we use the following strategy. We start by computing the expected center of mass of the object as, 
$\hat{x} = \sum_{k=1}^{K} P(\mathcal{M}^{(k)}, \mu^{(k)}) \frac{\sum_{i\in \{1,\dots,n\}}  (\mathcal M^{(k)}_i) (p^i_{x,t},p^i_{y,t})}{\sum_{i\in \{1,\dots,n\}} \mathcal M^{(k)}_i}$
We then find a cell $i$ from the object's outer envelope as
$i = \arg\max_{i\in \{1,\dots,n\}} \frac{ \big(\hat{x} - x_{target}\big)\big(  (p^i_{x,t},p^i_{y,t}) - \hat{x} \big)^T }{\| \hat{x} - x_{target}  \|_2   \|(p^i_{x,t},p^i_{y,t}) - \hat{x} \|_2 } $. In other terms, $i$ is the outer cell that is most aligned to the axis $\hat{x} - x_{target}$. The direction of the pushing force is always selected as the opposite of the surface normal at the contact point. The next step consists in simulating the pushing force and predicting the next state of the object (line 12) according to each (mass, friction) model in $\Theta$. The expected gap (or cost) of the simulated action is the weighted average of the distances between predicted poses, according to different models, and the target pose $x_{target}$, wherein the weights are the probabilities of the various models. This process is repeated by perturbing the location of the contact point, and retaining at the end the one that yields the minimum gap from the target pose. 
\subsection{Application: Pre-Grasp Sliding Manipulation}
The proposed probabilistic model identification technique is used in Algorithm~\ref{algo} to perform pre-grasp sliding manipulation of unknown objects. The algorithm integrates the previous ones to solve the problem of pushing an unknown object to the edge of a table and grasping it safely from there. We assume that the goal region is known. First, we start by using Algorithm~\ref{identificationAlgo} to explore the object and identify a set $\Theta$ of possible (mass, friction) models and their probabilities $P$. We then search for a configuration in the goal region that satisfies two criteria: 1) the object can be grasped without colliding with the support surface, and 2) the object is stable with a sufficiently high probability. To this end, a large number of object poses in the goal region are sampled and verified. Sampled poses that do not allow for a collision-free form-closure grasp are eliminated. Also, for each sampled pose, we use the identified mass models to predict in simulation the stability of the object. The failure probability is computed by adding together the probabilities of all mass models where the object is predict to fall from the table. Only goal configurations that ensure the stability of the object with a probability higher than $1-\epsilon$ are selected. The robust planning algorithm~\ref{policyAlgo} is then used to select a sequence of pushing actions to displace the object to the selected goal configuration. 

  \begin{algorithm}[h]
  {  \small
     \SetAlgoLined
\KwIn{Point cloud of an object's upper surface; Maximum mass and friction $m_{max},u_{max}\in \mathbb{R}$; Desired goal region $\mathcal G$; Failure probability threshold $\epsilon$\;}
Create a 3D shape $\mathcal S$ of the object by projecting its upper surface down on the support surface\; 
Decompose the 3D shape into a grid of $n$ small cuboids\;  
Let $x^g_0$ be the vector of ground-truth positions and rotations of the $n$ cuboids at time $0$; $\dot{x}^g_0 = \mathbf{0}$\;
Use Algorithm~\ref{identificationAlgo} to manipulate the object and identify  a set $\Theta$ of possible joint mass and friction models and their corresponding probabilities $P$\;
Sample a goal configuration ${x}^d \in \mathcal G$ where 1) a form-closure grasp of the object is available, and 2) the probability
that the object remains stable under gravity, according to the identified mass distribution, is higher than a threshold $1-\epsilon$\;
Use Algorithm~\ref{policyAlgo} to obtain a new sequence of forces $(F_{t})_{t=0}^{T_{\textrm{\tiny execution}}}$ to push the object to ${x}^d$\;
Execute the sequence $(F_{t})_{t=0}^{T_{\textrm{\tiny execution}}}$ with the robotic hand\;
Apply a form-closure grasp to pick up the object\;
}
\caption{Pre-Grasp Sliding Manipulation of Unknown Objects}
\label{algo}
\end{algorithm}

%% file: experiments.tex
\begin{figure*}[h]
\centering
\begin{tabular}{|c|c|c | c|c|c | c|c|c|}\hline
\multicolumn{3}{|c|}{Box} & 
\multicolumn{3}{c|}{Hammer} & 
\multicolumn{3}{c|}{Wrench}\\\hline
\multicolumn{3}{|c|}{
\includegraphics[width=0.13\textwidth]{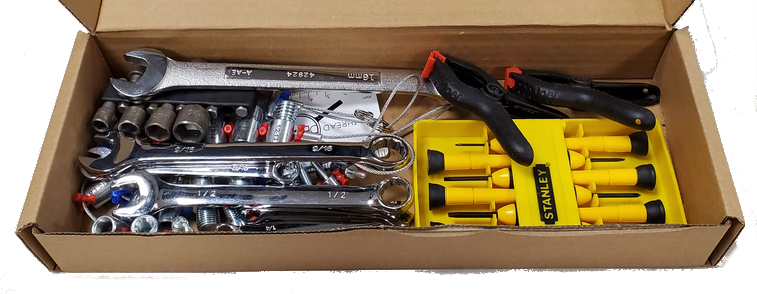} \quad
\includegraphics[width=0.13\textwidth]{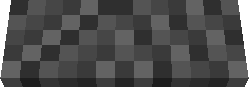}} & 
\multicolumn{3}{c|}{
\includegraphics[width=0.13\textwidth]{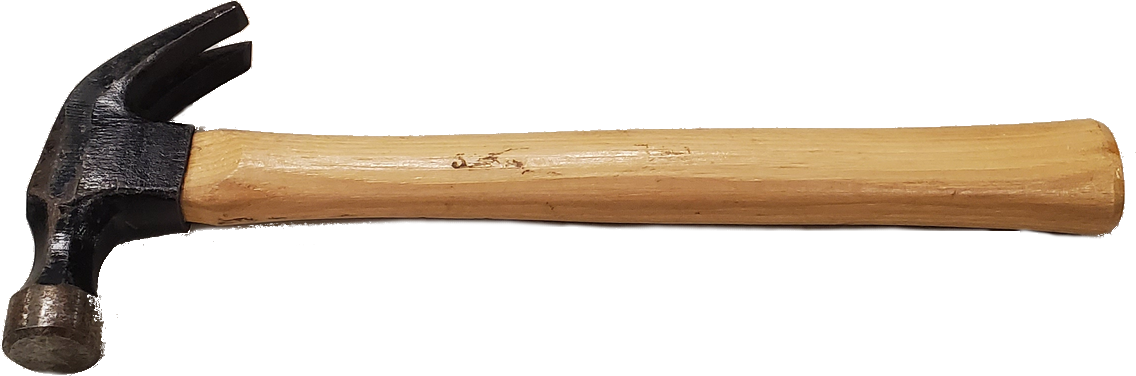}  \quad
\includegraphics[width=0.13\textwidth]{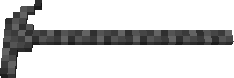}} &
\multicolumn{3}{c|}{
\includegraphics[width=0.13\textwidth]{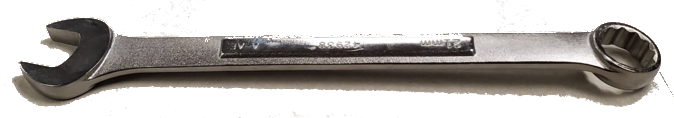}  \quad
\includegraphics[width=0.13\textwidth]{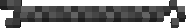}} \\\hline
\footnotesize{mass$\times$friction} & mass & friction &
\footnotesize{mass$\times$friction} & mass & friction &
\footnotesize{mass$\times$friction} & mass & friction \\\hline
\includegraphics[width=0.08\textwidth]{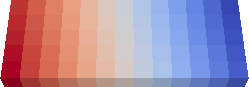} &
\includegraphics[width=0.08\textwidth]{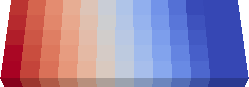} &
\includegraphics[width=0.08\textwidth]{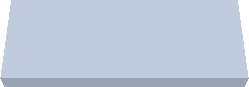} &
\includegraphics[width=0.08\textwidth]{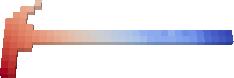} &
\includegraphics[width=0.08\textwidth]{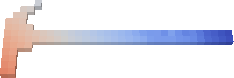} &
\includegraphics[width=0.08\textwidth]{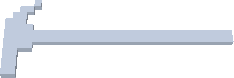} &
\includegraphics[width=0.08\textwidth]{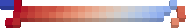} &
\includegraphics[width=0.08\textwidth]{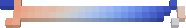} &
\includegraphics[width=0.08\textwidth]{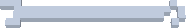} \\
\hline
\includegraphics[width=0.08\textwidth]{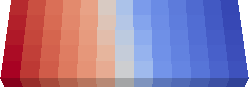} &
\includegraphics[width=0.08\textwidth]{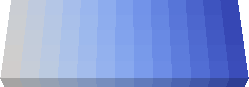} &
\includegraphics[width=0.08\textwidth]{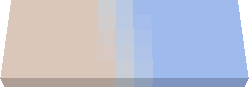} &
\includegraphics[width=0.08\textwidth]{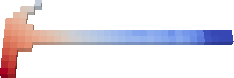} &
\includegraphics[width=0.08\textwidth]{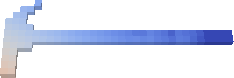} &
\includegraphics[width=0.08\textwidth]{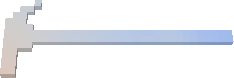} &
\includegraphics[width=0.08\textwidth]{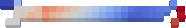} &
\includegraphics[width=0.08\textwidth]{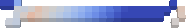} &
\includegraphics[width=0.08\textwidth]{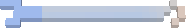} \\
\hline
\includegraphics[width=0.08\textwidth]{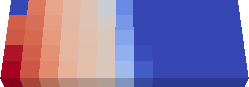} &
\includegraphics[width=0.08\textwidth]{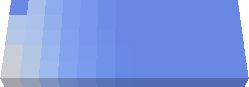} &
\includegraphics[width=0.08\textwidth]{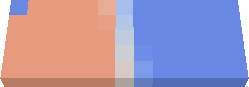} &
\includegraphics[width=0.08\textwidth]{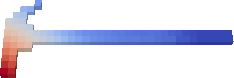} &
\includegraphics[width=0.08\textwidth]{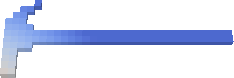} &
\includegraphics[width=0.08\textwidth]{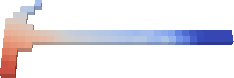}&
\includegraphics[width=0.08\textwidth]{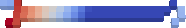} &
\includegraphics[width=0.08\textwidth]{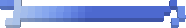} &
\includegraphics[width=0.08\textwidth]{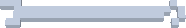}\\
\hline\hline
\multicolumn{3}{|c|}{Book} & 
\multicolumn{3}{c|}{Snack} & 
\multicolumn{3}{c|}{Spray Gun}\\\hline
\multicolumn{3}{|c|}{
\includegraphics[width=0.10\textwidth]{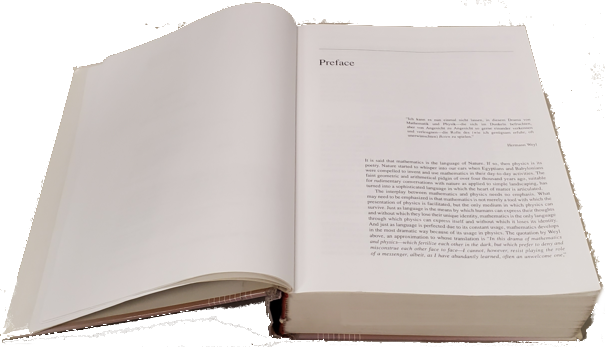} \quad
\includegraphics[width=0.10\textwidth]{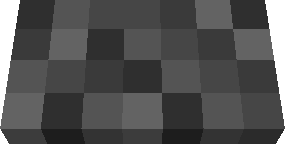}} & 
\multicolumn{3}{c|}{
\includegraphics[width=0.09\textwidth]{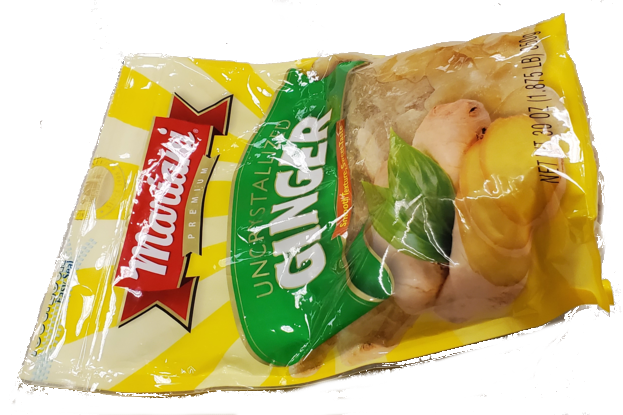}  \quad
\includegraphics[width=0.07\textwidth]{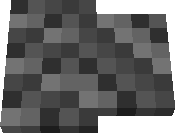}} &
\multicolumn{3}{c|}{
\includegraphics[width=0.13\textwidth]{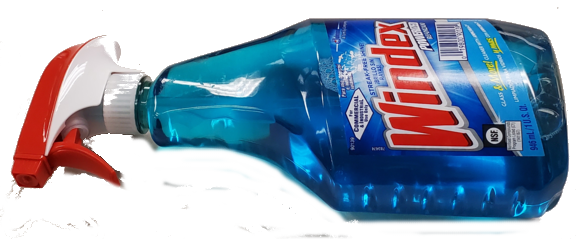}  \quad
\includegraphics[width=0.13\textwidth]{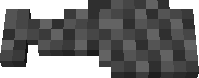}} \\\hline
\footnotesize{mass$\times$friction} & mass & friction &
\footnotesize{mass$\times$friction} & mass & friction &
\footnotesize{mass$\times$friction} & mass & friction \\\hline
\includegraphics[width=0.08\textwidth]{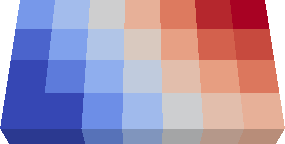} &
\includegraphics[width=0.08\textwidth]{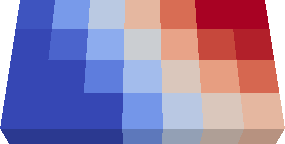} &
\includegraphics[width=0.08\textwidth]{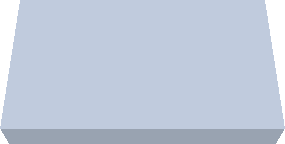} &
\includegraphics[width=0.07\textwidth]{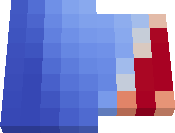} &
\includegraphics[width=0.07\textwidth]{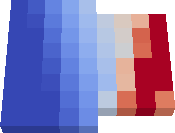} &
\includegraphics[width=0.07\textwidth]{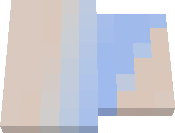} &
\includegraphics[width=0.08\textwidth]{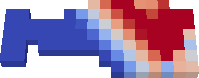} &
\includegraphics[width=0.08\textwidth]{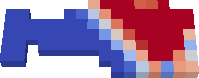} &
\includegraphics[width=0.08\textwidth]{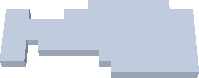} \\
\hline
\includegraphics[width=0.08\textwidth]{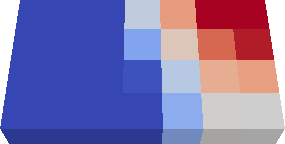} &
\includegraphics[width=0.08\textwidth]{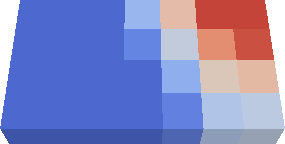} &
\includegraphics[width=0.08\textwidth]{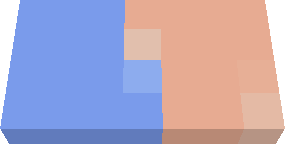} &
\includegraphics[width=0.07\textwidth]{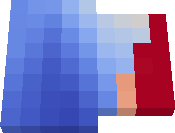} &
\includegraphics[width=0.07\textwidth]{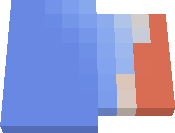} &
\includegraphics[width=0.07\textwidth]{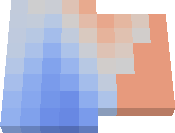} &
\includegraphics[width=0.08\textwidth]{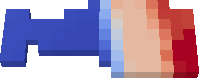} &
\includegraphics[width=0.08\textwidth]{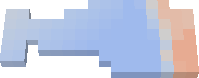} &
\includegraphics[width=0.08\textwidth]{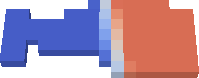} \\
\hline
\includegraphics[width=0.08\textwidth]{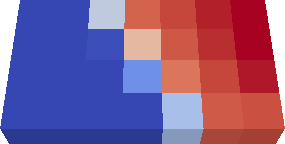} &
\includegraphics[width=0.08\textwidth]{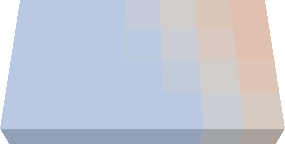} &
\includegraphics[width=0.08\textwidth]{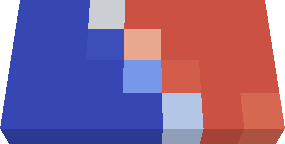} &
\includegraphics[width=0.07\textwidth]{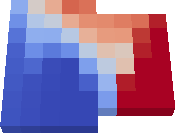} &
\includegraphics[width=0.07\textwidth]{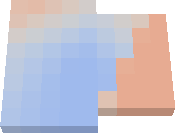} &
\includegraphics[width=0.07\textwidth]{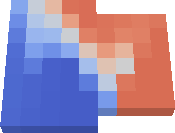}&
\includegraphics[width=0.08\textwidth]{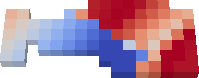} &
\includegraphics[width=0.08\textwidth]{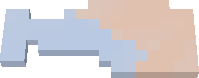} &
\includegraphics[width=0.08\textwidth]{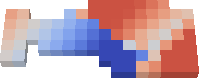}\\
\hline
\end{tabular}
\caption{Cuboid representations of real objects, along with samples from distribution $P$ of joint models returned by Algorithm~\ref{identificationAlgo} using a real robot to collect data. 
Red indicates higher values. Note how different combinations of mass and friction maps can result in the same frictional forces.}
\label{fig:massfric}
\end{figure*}

\section{Experiments}
We report here the results of experiments on pre-grasp sliding manipulation both in simulation and using a real robot and objects to evaluate the proposed approach. A video of the experiments is attached as a supplementary material.

\subsection{Experimental Setup}
The experiments are performed on both simulated and real robot and six objects: a book, a hammer, a box, a wrench, a snack, and a spray gun.
The simulation experiments are performed using the physics engine \textit{Bullet} and models of the robot and objects.
In the real setup, we used a \textit{Kuka} robot with a \textit{Robotiq} 3-finger hand and an RGB-D camera. % on the hand.
%The object pose is observed using a tag.

\subsection{Compared Methods}
The proposed algorithm is compared against the following methods. \texttt{Planning with Uniform Mass Distribution} plans each object's trajectory assuming it has a uniform mass and friction. Thus, each cuboid in the object has the same mass and coefficient of friction as the other cuboids.
\texttt{Deterministic Planning with Identified Mass Distribution} uses Algorithm~\ref{identificationAlgo} to learn the material properties  from the same data as our method, but samples one model of mass and friction $(\mathcal{M}^{(k)}, \mu^{(k)})\sim P$, and plans the object's entire trajectory based on the sampled model. All the results of this method are averaged over ten samples of $(\mathcal{M}^{(k)}, \mu^{(k)})\sim P$. 
Note that each of the sampled models is already locally optimal and yields accurate predictions of the sliding motions.
Because ground-truth mass and friction are known in the simulation experiments, we also compare the proposed algorithm with \texttt{Planning with Ground-truth Mass Distribution} which provides an upper bound.

\subsection{Model Identification Results with Real Robot and Objects}
Figure~\ref{fig:massfric} shows qualitative results obtained from Algorithm~\ref{identificationAlgo} and five pushing actions per object.
Each object is represented by a set of cuboids shown in the first rows. The number of cells per object varies from 28 to 88 depending on the size of the object.
%We obtained $9$ different mass and friction values by setting different upper bounds for mass and friction during model identification. 
Notice that models of the same object have different mass and friction maps, however, they produce similar frictional forces given by the product of the mass with the friction at each cell.
This ambiguity justifies the need for probabilistic reasoning about the models to avoid pushing the object into unbalanced configurations. 
%The products of identified mass and friction values in the first column in Figure~\ref{fig:massfric} are similar each other while mass and friction values in the second and third column are variant.

\begin{figure}[h]
    \centering
    \includegraphics[width=0.47\textwidth]{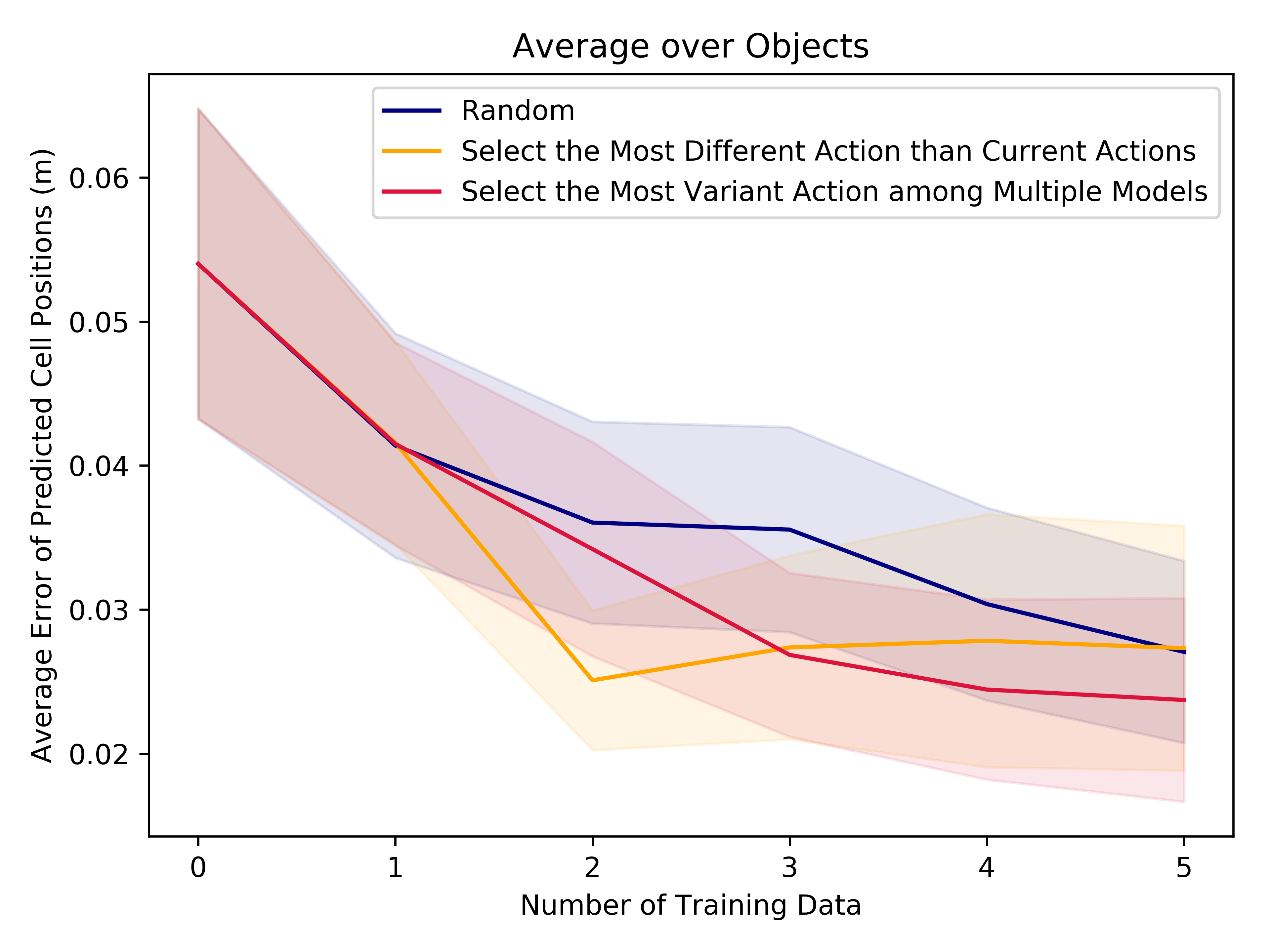}
    \caption{The average predicted cell position error (in meters) as a function of the number of training data. The results indicate that active action selection does not offer any clear advantage over a random strategy in this task.}
    \label{fig:res_ntrain}
\end{figure}

\begin{figure}[h]
    \centering
    \includegraphics[width=0.47\textwidth]{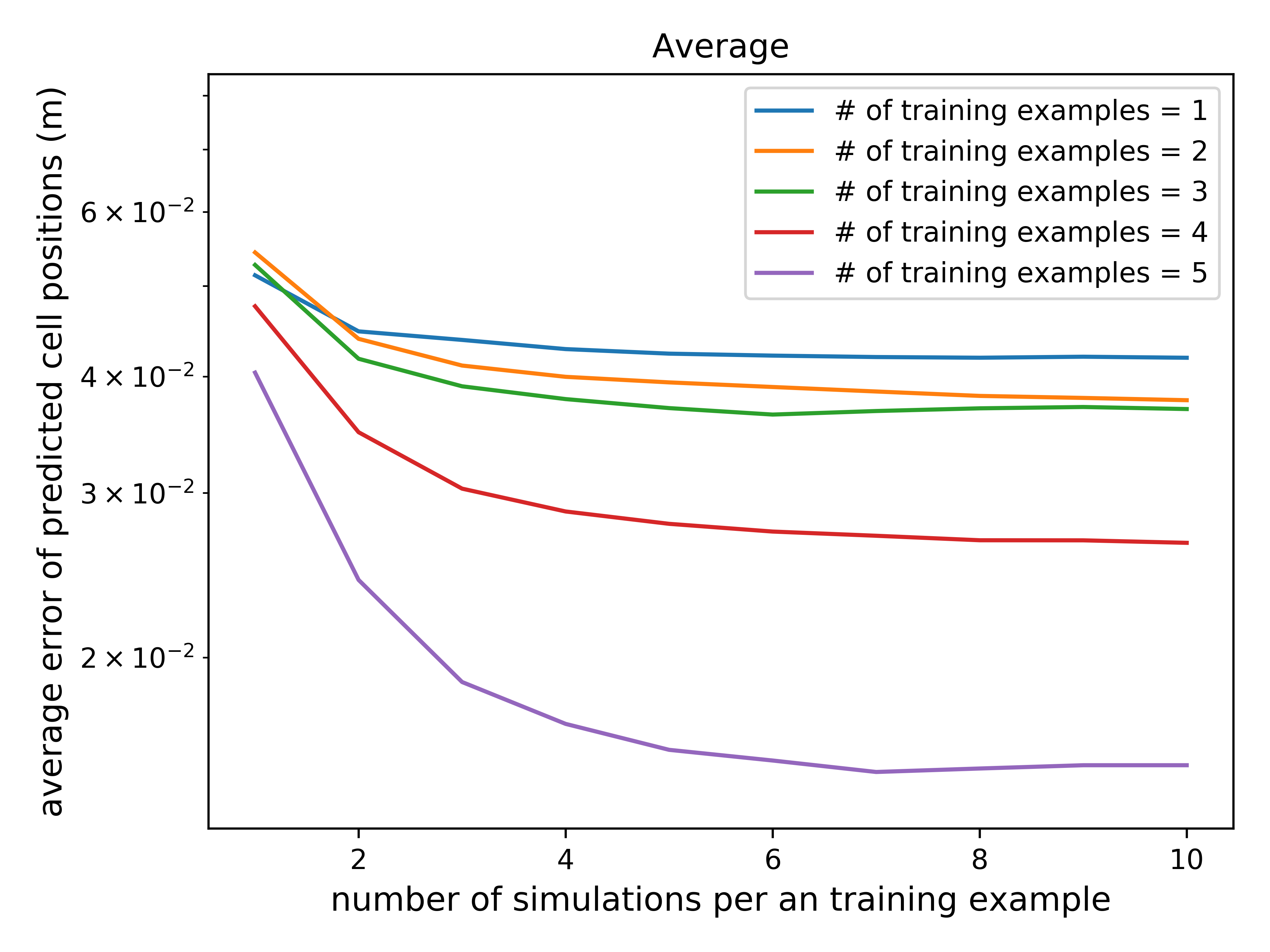}
    \caption{The average predicted cell position error of real objects (in meters) with different numbers of training examples, given the same upper bounds for the mass and friction coefficients as the ones used in~\cite{ChangkyuRSS2020}.}
    \label{fig:res_ntrain_aver}
\end{figure}

\subsection{Balance Prediction with Real Objects}
\label{sec:res_goal_selection}
\begin{figure}[h]
\includegraphics[width=0.47\textwidth]{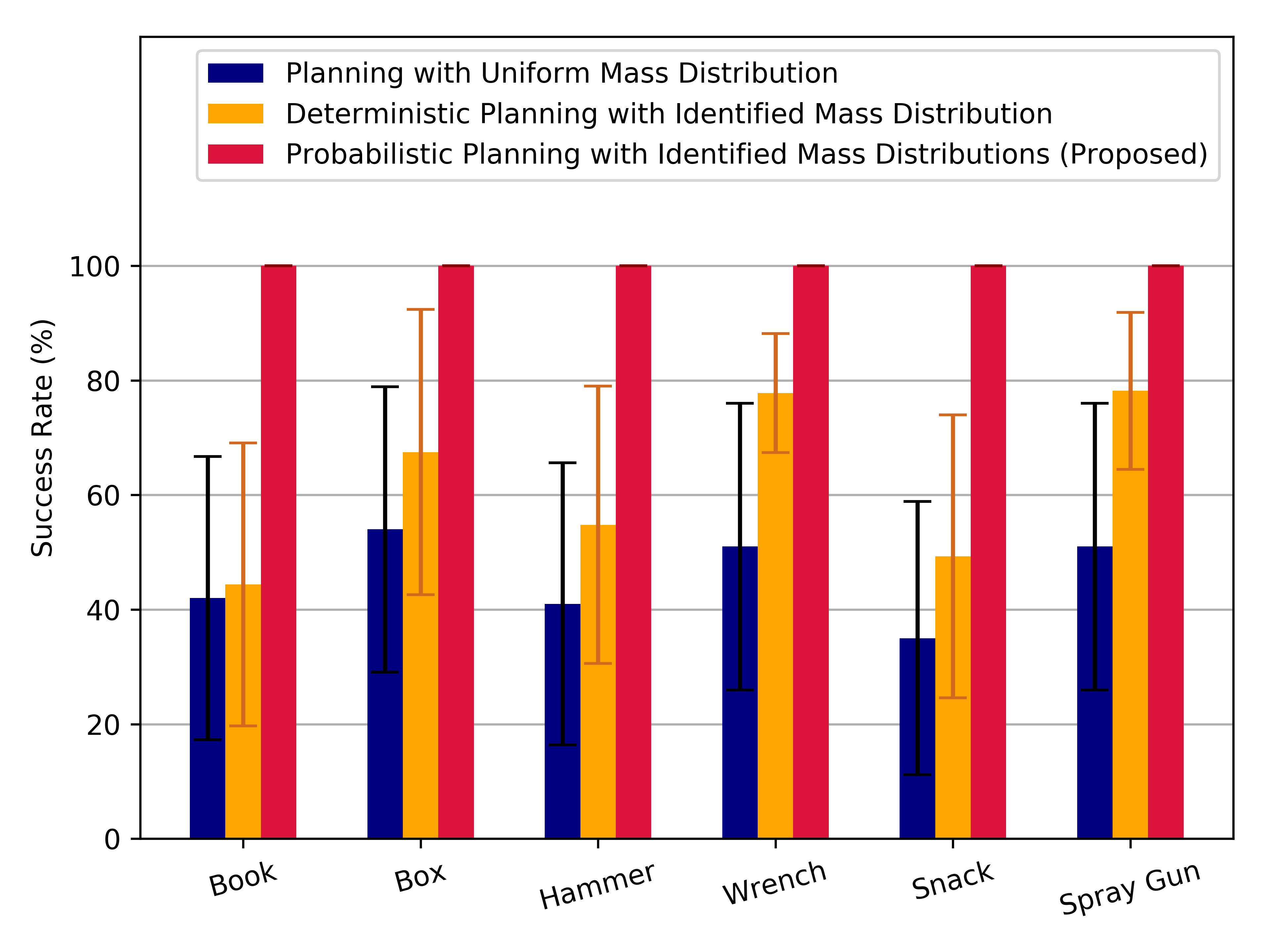}
\caption{Percentages of stable goal configurations selected by different methods using real robot and objects}
\label{fig:res_goal_selection}
\end{figure}

Figure~\ref{fig:res_goal_selection}  evaluates the accuracy of the proposed algorithm in selecting goal configurations based on the stability of the object.
We manually placed each of the six real objects in $10$ random poses on the edge of a round table where the form-closure grasp of the object is available.
And we recorded whether the objects fell or stayed on the table in each pose.
\texttt{Planning with Uniform Mass Distribution} and \texttt{Deterministic Planning with Identified Mass Distribution} returned any pose that is predicted to be stable in simulation using the identified or provided models. 
The proposed algorithm returns any pose where the object is predicted to remain balanced with a probability higher than the threshold of $0.90$.
The reported results are averaged over $100$ independent trials per object. The proposed probabilistic method obtained the highest success rate, 
because it considers many possible mass models, with different probabilities, and selects the safest goal poses. The other methods select one mass model per trial for predicting the balance of the object. 
Again, note that all the mass and friction models here are obtained from the same identification process, and they are all local minima of the loss function with virtually similar objective values.

\subsection{Pre-grasp Sliding Manipulation}
\subsubsection{Simulation Experiments}

\begin{figure}[h]
\centering
\begin{tabular}{c}
\includegraphics[width=0.47\textwidth]{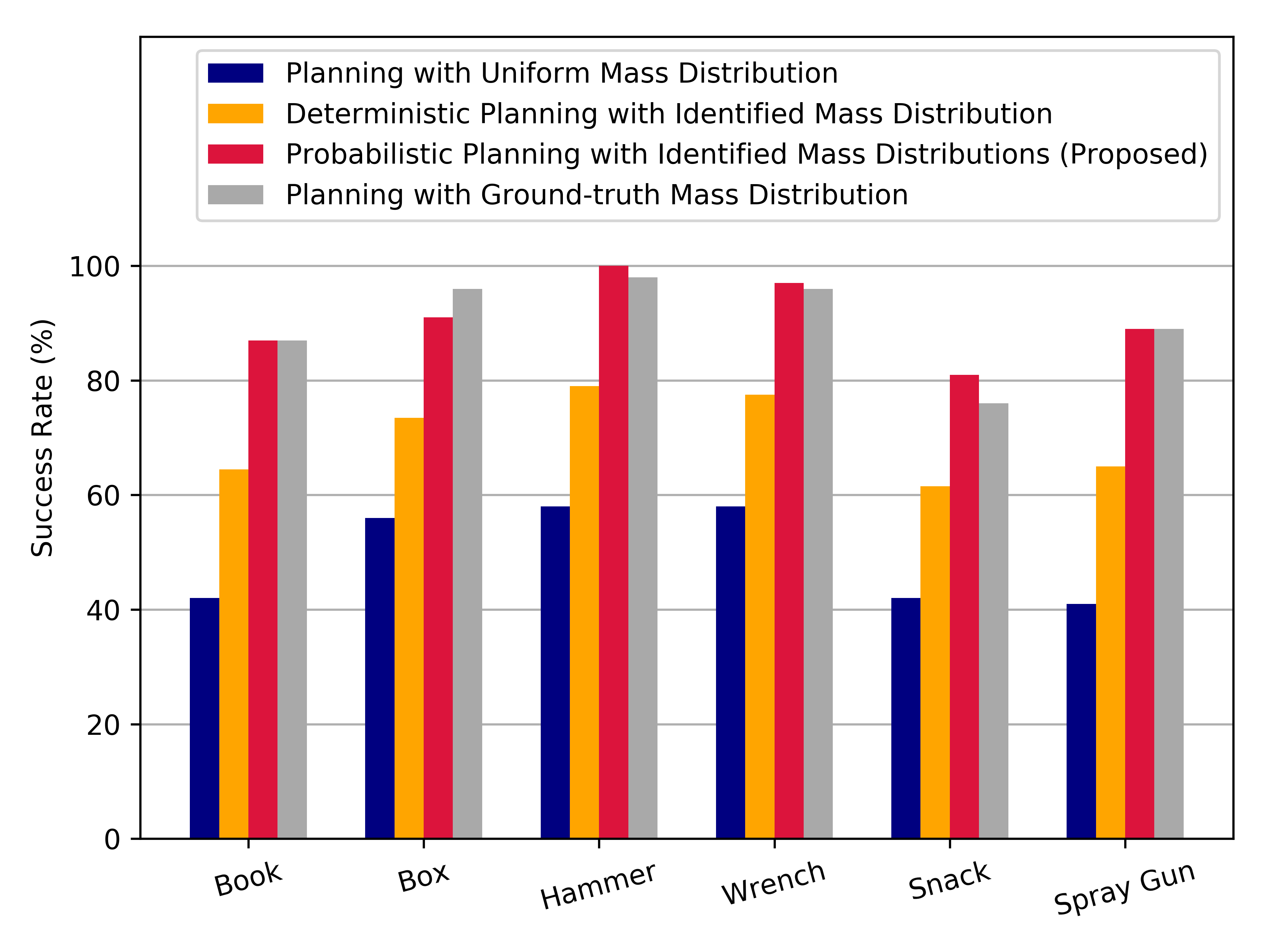}\\
{\small (a) Percentage of successful grasps} \\\\
\begin{tabular}{|l|c|}\hline
Method & \# of Actions\\\hline
Planning with Uniform Mass & $6.05~(\pm 1.76)$ \\
Deterministic Planning & $5.69~(\pm 1.96)$ \\
Probabilistic Planning (Proposed)& $6.27~(\pm 1.82)$ \\
Planning with Ground-truth Mass & $5.46~(\pm 1.44)$\\\hline
\end{tabular}
\\\\
{\small (b) Average number of executed actions}\\
\end{tabular}
\caption{Simulation results on pre-grasp sliding manipulation}
\label{fig:res_plan_sim}
\end{figure}

%Nine different mass and friction distributions were identified for each object, and these models are used in the proposed method. The deterministic planning method randomly chose one of the these models and used it during planning.
Each of the objects is placed in $100$ random rotations at the center of a round table with a radius of $0.5m$. After identifying the models of each object from five pushing actions, the goal is to select
a sequence of pushing actions that displace the object into graspable goal configurations at the edge of the table.
To make the experiments more challenging, we added a control noise to the simulator so that planning does not perfectly match with execution.
Figure~\ref{fig:res_plan_sim} (a) shows that in all of the $100$ trials with random initial object poses, the proposed method successfully placed the object in graspable and balanced poses.
\textit{Planning with Uniform Model Identification} failed mostly because it chose unstable poses as a goal due to the fact that most objects do not have a uniform mass.
\textit{Deterministic Planning with Identified Mass Distribution} also dropped the object when it chose inaccurate mass maps, which results in selecting unstable goal poses.
As shown in Figure~\ref{fig:res_ntrain_aver}, all the identified mass models perfectly predicted the motions of the objects while they were entirely on the surface because  they were used jointly with inaccurate friction maps that compensate for the mass in the product 
mass$\times$friction, as shown in Figure~\ref{fig:massfric}. 
The proposed method achieves the highest success rate, with a slightly higher number of pushing actions than other methods, as shown in Figure~\ref{fig:res_plan_sim} (b).

\subsubsection{Real Robot Experiments}

\begin{figure}[h]
\includegraphics[width=0.47\textwidth]{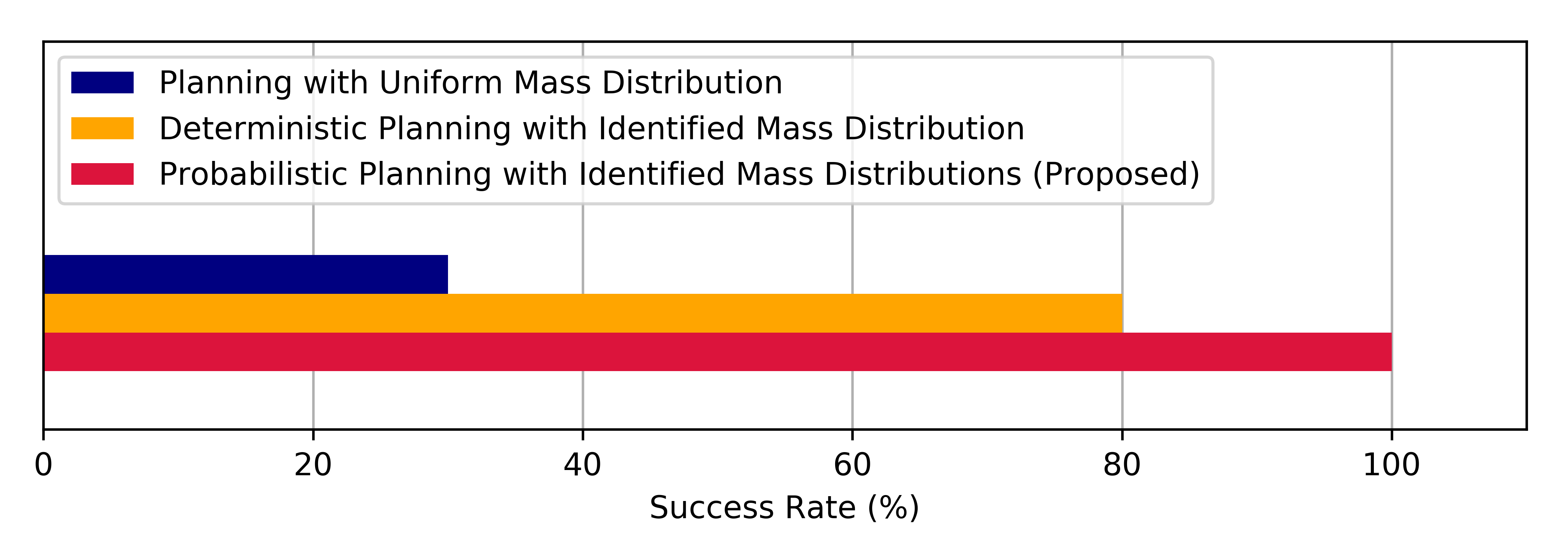}
\caption{Real robot results on pre-grasp sliding manipulation}
\label{fig:res_plan_real}
\end{figure}

We repeated the same experiment as the previous one using the real robotic setup with a hammer set in $10$ different initial poses on the table.
The \textit{Kuka} robot repeatedly observes the object's pose on the table and pushes the object toward pre-grasp poses, using Algorithm~\ref{algo}.
%We defined a number of possible graspable object poses hanging on the edge of the table, which guarantees enough space for robot to grasp the object.
Once the object reaches a goal pose, the robot grasps the object by inserting its fingers around it and aligning the approaching direction to the surface normal of the contact point. 
Figure~\ref{fig:res_plan_real} shows the percentage of successful grasps resulting from the entire process of Algorithm~\ref{algo}.
The lower success rate of \textit{Planning with Uniform Mass Distribution} and \textit{Deterministic Planning with identified Mass Distribution} mostly comes from unstable goal pose selections.
The \textit{Uniform Mass Distribution} shows much lower success rate than simulation results.
This is because reachable object poses with the real robot are more limited than in simulation, and plans obtained from inaccurate mass distributions lead into pushing the object out of the workspace. 
The proposed method was able to avoid those situations by considering many plausible models simultaneously and choosing actions that are most guaranteed to succeed.
\section{Conclusion}
Manipulating non-uniform objects safely requires accurate models of mass and friction maps. In this work, we proposed a new method for identifying such models automatically from data collected by using a robotic manipulator.
When the data is limited and collected from quasi-static motions, it is not possible to disentangle mass from friction. Therefore, the proposed approach returns a probability distribution over several plausible models. We have also 
shown that the inferred probability distribution can be used to plan side-pushing actions that displace an object into the edge of  a support surface while keeping it balanced. The object can then be grasped easily from the edge.
Future work includes investigating other probabilistic mass and friction representations, as well as inferring models of objects in highly cluttered scenes where contact and collisions between different unknown objects may occur.